\documentclass[runningheads]{llncs}
\usepackage{csquotes}
\usepackage{relsize}
\usepackage{adjustbox}
\usepackage{tikz}
\usepackage{pgfplots}
\usepackage{hyperref}
\usepackage{breakurl}
\usepackage{flushend} 
\usepackage{caption}

\usepackage{graphicx}
\usepackage{alltt,amsmath,amssymb}
\usepackage{csquotes}
\usepackage{cite}
\usepackage{enumitem}
\usepackage{float}
\floatstyle{ruled}
\newfloat{program}{t}{lop}
\floatname{program}{Program}
\usepackage{subcaption}

\newcommand{\etal}{\textit{et al.}}
\newcommand{\smartinduct}{\texttt{smart\_induct}}
\newcommand{\induct}{the \texttt{induct} tactic}
\newcommand{\lifter}{\texttt{LiFtEr}}

\newcommand{\indrule}{\texttt{rule}}
\newcommand{\term}{\texttt{term}}

\begin{document}
\title{Smart Induction for Isabelle/HOL \\(System Description)
\thanks{
This work was supported by the European Regional Development Fund under the project 
AI \& Reasoning (reg. no.CZ.02.1.01/0.0/0.0/15\_003/0000466).}
}
%
%
\author{Yutaka Nagashima
\inst{1}\inst{2}
\orcidID{0000-0001-6693-5325}}
\authorrunning{Y. Nagashima}
%
\institute{
Czech Technical University in Prague, Prague, Czech Republic
\email{Yutaka.Nagashima@cvut.cz}
\and
University of Innsbruck, Innsbruck, Austria}
\maketitle              
\begin{abstract}
Proof assistants
offer tactics to facilitate inductive proofs. 
However, 
it still requires human ingenuity to decide 
what arguments to pass to these tactics. 
To automate this process, 
we present \smartinduct{} for Isabelle/HOL.
Given an inductive problem in any problem domain,
\smartinduct{} lists promising arguments for the \texttt{induct} tactic 
without relying on a search. 
Our evaluation demonstrated \smartinduct{} produces valuable recommendations across problem domains.
\end{abstract}
\section{Induction}
Given the following two simple reverse functions defined in Isabelle/HOL \cite{isabelle},
how do you prove their equivalence \cite{concrete_semantics} ?

\begin{verbatim}
primrec rev::"'a list =>'a list" where
  "rev  []      = []"
| "rev (x # xs) = rev xs @ [x]"
fun itrev::"'a list =>'a list =>'a list" where
  "itrev  []    ys = ys"
| "itrev (x#xs) ys = itrev xs (x#ys)"
lemma "itrev xs ys = rev xs @ ys"    
\end{verbatim}
\noindent
where \verb|#| is the list constructor, 
and \verb|@| appends two lists.
%
Using \induct{} of Isabelle/HOL, we can prove this inductive problem in multiple ways:
%
\begin{verbatim}
lemma prf1: "itrev xs ys = rev xs @ ys"
  apply(induct xs arbitrary: ys) by auto
lemma prf2: "itrev xs ys = rev xs @ ys"
  apply(induct xs ys rule:itrev.induct) by auto
\end{verbatim}

\noindent
\verb|prf1| applies structural induction on \verb|xs|
while generalizing \verb|ys| before applying induction
by passing \verb|ys| to the \texttt{arbitrary} field.
On the other hand, \verb|prf2| applies functional induction on \verb|itrev| by passing an auxiliary lemma, \verb|itrev.induct|,
to the \texttt{rule} field.

There are other lesser-known techniques to handle difficult inductive problems using \induct{}, 
and sometimes users have to 
develop useful auxiliary lemmas manually;
However, for most cases
the problem of how to apply induction 
boils down to the the following three questions:

\begin{itemize}[noitemsep, nolistsep]
    \item On which terms do we apply induction?
    \item Which variables do we generalize using the \texttt{arbitrary} field?
    \item Which rule do we use for functional induction using the \texttt{rule} field?
\end{itemize}

To answer these questions automatically, 
we developed a proof strategy language, \texttt{PSL} \cite{psl}.
Given an inductive problem,
\texttt{PSL} produces various combinations of induction arguments for \induct{}
and conducts an extensive proof search based on a given strategy.
If \texttt{PSL} completes a proof search, 
it identifies the appropriate combination of arguments for the problem
and presents the combination to the user;
However, when the search space becomes enormous,
\texttt{PSL} cannot complete a search within a realistic timeout
and fails to provide any recommendation,
even if \texttt{PSL} produces the right combination of induction arguments.
For further automation of proof by induction,
we need a tool that satisfies the following two criteria:

\begin{itemize}[noitemsep, nolistsep]
  \item The tool suggests right induction arguments without resorting to a search.
  \item The tool suggests right induction arguments for any inductive problems.
\end{itemize}

In this paper we present \smartinduct{}, a recommendation tool that addresses these criteria.
\smartinduct{} is available at GitHub \cite{GitHub}
together with our running example and the evaluation files discussed in Section \ref{s:eval}.
The seamless integration of \smartinduct{} into Isabelle's ecosystem 
made \smartinduct{} easy to install and easy to use.
The implementation of \smartinduct{} is specific to Isabelle/HOL;
However, the underlying concept is transferable to other 
tactic based proof assistants including HOL4 \cite{hol4}, Coq \cite{coq}, and Lean \cite{Lean}. 

To the best of our knowledge \smartinduct{} is the first recommendation
tool that analyzes the syntactic structures of proof goals across problem domains and 
advise how to apply \induct{} without resorting to a search.




\begin{figure}[t]
\centering
  \includegraphics[width=0.7\textwidth]{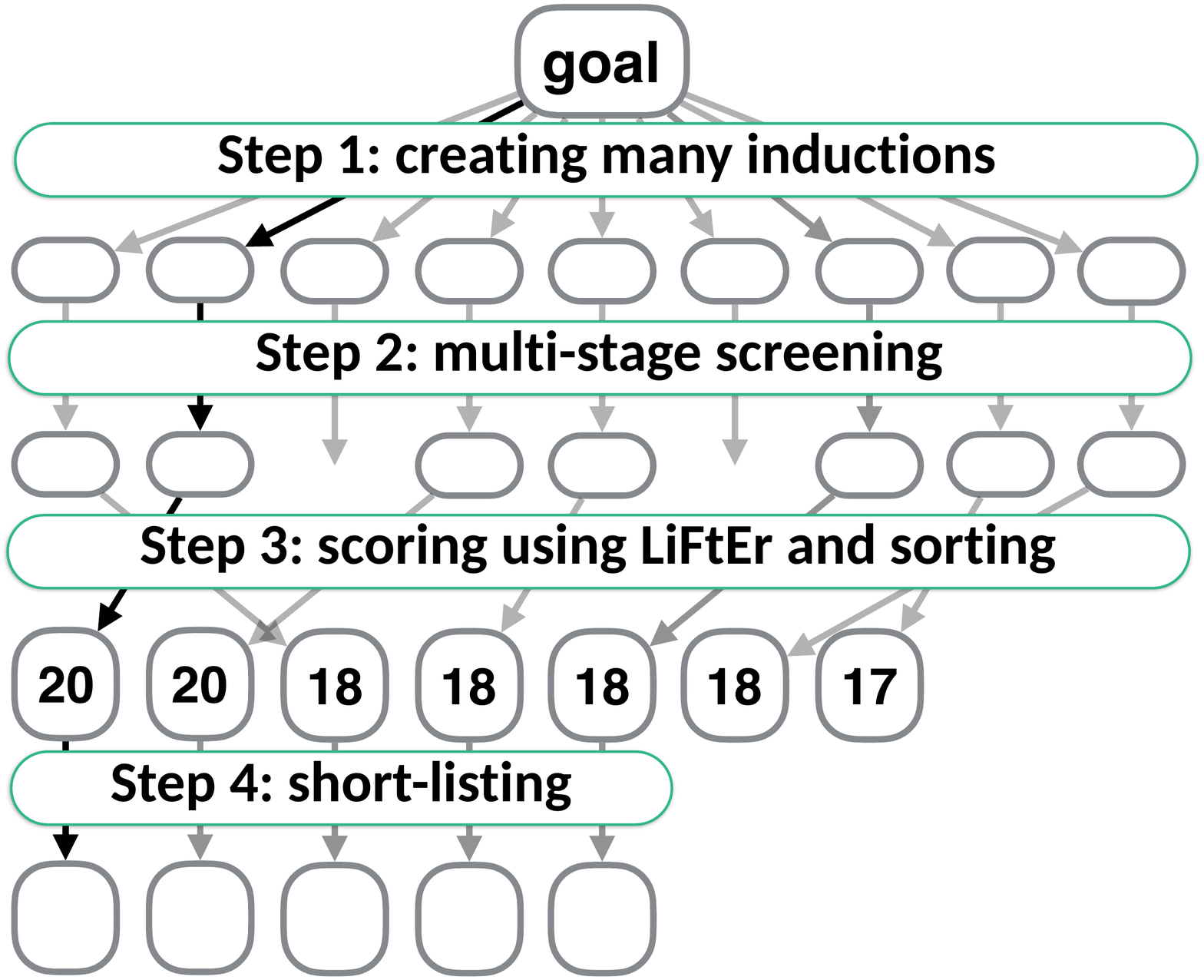}
  \caption{The Workflow of \smartinduct.}
  \label{fig:smartinduct}
\end{figure}

\section{\smartinduct{}: the System Description}
Fig. \ref{fig:smartinduct} illustrates the following internal workflow of \smartinduct{}.
When invoked by a user, 
the first step produces many variants of \induct{} with different combinations of arguments.
Secondly, the multi-stage screening step filters out less promising combinations induction arguments.
Thirdly, the scoring step evaluates each combination to a natural number using
logical feature extractors implemented in \texttt{LiFtEr} \cite{lifter}
and reorder the combinations based on their scores.
Lastly, the short-listing step takes the best 10 candidates and print them in the Output panel of Isabelle/jEdit.
In this section, we explore details of Step 1 to Step 3.

\subsection{Step 1: Creation of Many Induction Tactics.}\label{s:manyinduct}
\smartinduct{} first inspects the given proof goal 
and produces a number of combinations of arguments for \induct{} taking the following procedure:
\smartinduct{} collects variables and constants appearing in the goal.
If a constant has an associated induction rule,
\smartinduct{} also collects that rule from the underlying proof context,
Then, from these variables and induction rules,
\smartinduct{} produces a power set of combinations of arguments for \induct.
In our example, \smartinduct{} produces 40 combinations of induction arguments.

If the size of this power set is enormous,
we cannot store all the produced induction tactics in our machines.
Therefore, \smartinduct{} produces this set using a lazy sequence and 
takes only the first 10000 of them for further processing.

\subsection{Step 2: Multi-Stage Screening.}\label{s:screening}
10000 is still a large number, and 
feature extractors used in Step 3 often 
often involve nested traversals of nodes in the syntax tree
representing a proof goal, 
leading to high computational costs.
Fortunately, 
the application of \induct{} itself is not computationally expensive in many cases:
We can apply \induct{} to a proof goal and have intermediate sub-goals at a low cost.
Therefore, in Step 2, \smartinduct{} applies \induct{} to the given proof goal using the various combinations of arguments
from Step 1 and filter out some of them through the following two stages.

\paragraph{Stage 1.}
In the first stage, \smartinduct{} filters out those combinations of induction arguments,
with which Isabelle/HOL does not produce an intermediate goal. 
Since we have no known theoretical upper bound for the computational cost for \induct{},
we also filter out those combinations of arguments,
with which \induct{} does not return a result within a pre-defined timeout.
In our running example, this stage filters out 8 combinations out of 40.

\paragraph{Stage 2.}
Taking the results from the previous stage,
Stage 2 scans both the original goal and the newly introduced intermediate sub-goals at the same time
to further filter out less promising combinations.
More concretely,
this stage filters out all combinations of arguments
if they satisfy any of the following conditions:

\begin{itemize}[noitemsep, nolistsep]
    \item Some of newly introduced sub-goals are identical to each other.
    \item All newly introduced sub-goals contain the original first sub-goal as their sub-term
even though there was no locally introduced assumptions.
    \item A newly introduced sub-goal contains a schematic variable even though the original first sub-goal did not contain a schematic variable.
\end{itemize}

\noindent
In our example, Stage 2 filters out 4 combinations out of 32.
Note that these tests on the original goal and resulting sub-goals
do not involve nested traversals of nodes in the syntax tree representing goals.
For this reason,
the computational cost of this stage is often smaller than that of Step 3.

\subsection{Step 3: Scoring Induction Arguments using \texttt{LiFtEr}.}

Step 3 carefully investigates the remaining candidates using heuristics implemented in \texttt{LiFtEr} \cite{lifter}.
\texttt{LiFtEr} is a domain-specific language to encode induction heuristics
in a style independent of problem domains.
Given a proof goal and combination of induction arguments,
the \lifter{} interpreter mechanically checks 
if the combination is appropriate for the goal
in terms of a heuristic written in \lifter.
The interpreter returns \texttt{True} if the combination is compatible with the heuristic
and \texttt{False} if not.
We illustrated the details of \lifter{} in our previous work \cite{lifter}
with many examples.
In this paper,
we focus on the essence of \lifter{} and show one example heuristic used in \smartinduct.

\lifter{} supports four types of variables:
natural numbers, induction rules, terms, and term occurrences.
An induction rule is an auxiliary lemma passed to the \indrule{} field of \induct{}.
The domain of terms is the set of all sub-terms appearing in a given goal.
The logical connectives ($\lor$, $\land$, $\rightarrow$, and $\neg$)
correspond to the connectives in the classical logic.
\lifter{} offers atomic assertions, such as 
\texttt{is\_rule\_of}, to examine the property of each atomic term.
Quantifiers bring the the power of abstraction to \lifter{}, which
allows \lifter{} users to encode induction heuristics
that can transcend problem domains.
Quantification over \term{} can be restricted to the induction terms used in \induct.

We encoded 20 heuristics in \lifter{} for \smartinduct.
Some of them examine a combination of induction arguments in terms of functional induction,
whereas others check the combination for structural induction or rule inversion. 
Program \ref{p:example}, for example, 
encodes a heuristic for functional induction.
In English this heuristic reads as follows:

\begin{program*}[!t]
\begin{alltt}
  \(\exists r1\) : rule. True
\(\rightarrow \)
  \(\exists r1\) : rule.
    \(\exists t1\) : term.
      \(\exists to1\) : term_occurrence \(\in t1\) : term.
          \(r1\) is_rule_of \(to1\)
        \(\land\)
          \(\forall t2\) : term \(\in\) induction_term.
            \(\exists to2\) : term_occurrence \(\in t2\) : term.
              \(\exists n\) : number.
                  is_nth_argument_of (\(to2\), \(n\), \(to1\))
                \(\land\)
                  \(t2\) is_nth_induction_term \(n\)
\end{alltt}
\caption{A \lifter{} Heuristic used in \smartinduct{}.}
\label{p:example}
\end{program*}

\begin{displayquote}
if there exists a rule, $r1$, 
in the \texttt{rule} field of the \verb|induct| tactic,
then there exists a term $t1$ with an occurrence $to1$, 
such that
$r1$ is derived by Isabelle when defining $t1$, and
for all induction terms $t2$,
there exists an occurrence $to2$ of $t2$ such that,
there exists a number $n$, such that
$to2$ is the $n$th argument of $to1$ and that
$t2$ is the $n$th induction terms passed to the \verb|induct| tactic.
\end{displayquote}

If we apply this heuristic to our running example, \texttt{prf2},
the \lifter{} interpreter returns \texttt{True}:
there is an argument, \texttt{itrev.induct}, in the \verb|rule| field,
and the occurrence of its related term, \verb|itrev|, in the proof goal takes
all the induction terms, \verb|xs| and \verb|ys|, as its arguments in the same order.

Attentive readers may have noticed that Program \ref{p:example} is independent of
any types or constants specific to \texttt{prf2}.
In stead of handling specific constructs explicitly,
Program \ref{p:example} analyzes the structure of the goal with respect to 
the arguments passed to \induct{}
in an abstract way using quantified variables and logical connectives.
This power of abstraction let \smartinduct{} evaluate whether 
a given combination of arguments to \induct{} is appropriate for 
a user-defined proof goal consisting of user-defined types and constants,
even though such constructs are not available to the \smartinduct{} developers.
In fact, none of the 20 heuristics relies on constructs specific to any problem domain.

In Step 3,
\smartinduct{} applies these 20 heuristics to the results from Step 2.
For each heuristic, \smartinduct{} gives one point to each combination of \texttt{induct} arguments
if the \lifter{} interpreter returns \texttt{True} for that combination.
Then, \smartinduct{} reorder these combinations based on their scores
to present the most promising combinations to the user in Step 4.

\section{Evaluation}\label{s:eval}
In general it is not possible to measure
if a combination of induction arguments is correct for a goal.
Therefore, we evaluated trustworthiness of \smartinduct{}'s recommendations
using \textit{coincident rates}:
We counted how often its recommendation coincides with the choices
of Isabelle experts.

\begin{table}[t]
\caption{Coincidence Rates of \smartinduct{}.}
\label{table:eval}
\begin{center}
\begin{tabular}{c c c c c c c c}
\hline\noalign{\smallskip}
theory & total & ~top\_1~ & ~top\_3~ & ~top\_5~ & ~top\_10\\
\hline
\noalign{\smallskip}
\verb|DFS|              & 10  & ~6~(60\%)~    & ~9~(90\%)~    & ~9~(90\%)~   & ~9~(90\%)~\\
\verb|Nearest_Neighbors|  & 16  & ~3~(19\%)~    & ~4~(25\%)~    & ~7~(44\%)~   & ~12~(75\%)~\\
\verb|RST_RBT|          & 24  & ~24~(100\%)~  & ~24~~(100\%)~ & ~24~(100\%)~ & ~24~(100\%)~\\
sum                     & 50  & ~33~(65\%)~ & ~37~(74\%)~ & ~40~(80\%)~ & ~45~(90\%)~\\
\hline
\end{tabular}
\end{center}
\end{table}

Table \ref{table:eval} shows the coincidence rates of \smartinduct{}
for three theory files
about different problem domains written by different researchers:
\texttt{DFS} is a part of the formalisation of depth-first search \cite{dfs},
\texttt{Nearest\_Neighbors} is from the foramlisation of multi-dimensional binary search trees \cite{kd_tree},
and \texttt{RST\_RBT} is from that of priority search tree \cite{priority_search_tree}.

The column named ``total'' shows 
the total number of proofs by induction in each file.
The four columns titled with ``top\_n'' show
for how many proofs by induction in each file
the proof authors' choice of induction arguments coincides with 
one of the n most promising recommendations from \smartinduct{}.
For example, top\_3 for \verb|DFS| indicates
if \smartinduct{} recommends 3 most promising combinations for each goal,
the proof author used one of the three recommended combinations
for 90\% of proofs by induction in the file.
Note that we often have multiple equally valid combinations of induction arguments for a given proof goal: 
Our running example has two proofs, \texttt{prf1} and \texttt{prf2},
and both of them are appropriate to prove this equivalence theorem.
Therefore, we should regard a coincidence rate as a conservative estimate of true success rate.

A quick glance over Table \ref{table:eval} would give the impression that
\smartinduct{}'s performance depends heavily on problem domains:
\smartinduct{} demonstrated the perfect result for \texttt{RST\_RBT},
whereas for \texttt{Nearest\_Neighbor} 
the coincidence rate remains at 44\% for top\_5.

However, a closer investigation of the results reveal that
the difference of performance comes from the style of induction
rather than domain specific items such as the types or constructs appearing in goals:
In \texttt{RST\_RBT}, all 24 proofs by induction are functional inductions,
whereas \texttt{Nearest\_Neighbor} has only 5 functional inductions out of 16.
As Table \ref{table:nearest_neibors} in Appendix shows
if we focus on proofs by functional induction
the coincidence rate rises to 60\% for ``top\_1'' and 80\% for ``top\_3''
for \texttt{Nearest\_Neighbor}.

Furthermore, Table \ref{table:nearest_neibors} also suggests that
\smartinduct{} has relatively low coincidence rates for structural induction
because \smartinduct{} is not able to predict which variables to generalize
using the \verb|arbitrary| field:
Since structural induction tends to involve generalization of variables 
more often than functional induction does,
\smartinduct{} struggles to predict 
the choice of experts for structural induction.

In Table \ref{table:only_on}
we computed the coincidence rates for \texttt{Nearest\_Neighbor} again based on a different criterion:
This time we ignored the \texttt{rule} and \texttt{arbitrary} fields and
took only induction terms into consideration.
The large discrepancies between the numbers in Table \ref{table:eval}
and those in Table \ref{table:only_on} indicate that
even for structural inductions
\smartinduct{} is often able to predict on which variables experts apply induction
but fails to predict which variables to generalize.

\begin{table}[t]
\caption{Coincidence Rates of \smartinduct{} Based Only on Induction Terms.}
\label{table:only_on}
\begin{center}
\begin{tabular}{c c c c c c c c}
\hline\noalign{\smallskip}
theory & total & ~top\_1~ & ~top\_3~ & ~top\_5~ & ~top\_10\\
\hline
\noalign{\smallskip}
\verb|Nearest_Neighbors|  & 16  & ~5~(31\%)~    & ~12~(75\%)~    & ~15~(94\%)~   & ~15~(94\%)~\\
\hline
\end{tabular}
\end{center}
\end{table}

The limited performance in predicting experts' use of the \texttt{arbitrary} field
stems from \lifter's limited capability to examine semantic information of proof goals.
Even though 
\lifter{} offers quantifiers, logical connectives, and atomic assertions to 
analyze the syntactic structure of a goal in an abstract way,
\lifter{} does not offer enough supports to analyze the semantics of the goal.
For more accurate prediction of variable generalization,
\smartinduct{} needs a language to analyze
not only the structure of a goal itself but also
the structure of the definitions of types and constants appearing in the goal in an abstract way.

\section{Conclusion}

We presented \smartinduct{}, a recommendation tool for proof by induction in Isabelle/HOL.
Our evaluation showed
\smartinduct{}'s excellent performance in recommending how to apply functional induction
and identifying induction terms for structural induction,
even though recommendation of variable generalization remains as a challenging task.
It is still an open question how far we can improve \smartinduct{}
by combining it with search based systems \cite{psl,pgt}
and approaches based on evolutionary computation \cite{evoluaionary_isabelle} or
statistical machine learning \cite{meloid}.


\paragraph{Related Work.}
The most well-known approach for inductive problems is 
called the Boyer-Moore waterfall model \cite{waterfall}.
This approach was invented for a first-order logic on Common Lisp.
ACL2 \cite{acl2} is a commonly used waterfall model based prover.
When deciding how to apply induction,
ACL2 computes a score, called \textit{hitting ratio}, 
to estimate how good each induction scheme is for the term which it accounts for
and proceeds with the induction scheme with the highest hitting ratio \cite{acl_book, induction}.

Compared to the hitting ratio used in ACL2,
\smartinduct{} analyzes the structures of proof goals directly using \lifter{}.
While ACL2 produces many induction schemes and computes their hitting ratios,
\smartinduct{} does not directly produce induction schemes but analyzes the given proof goal,
the arguments passed to the \texttt{induct} method, 
and the emerging sub-goals.

Jiang \etal{} ran multiple waterfalls \cite{jiang} in HOL Light \cite{hollight}.
However, when deciding induction variables, they naively picked the first free variable with recursive type and 
left the selection of appropriate induction variables as future work.

Machine learning applications to tactic-based provers \cite{pamper,holist,tactictoe}
focus on selections of tactics rather than selections of terms as arguments to tactics,
even though the choice of induction arguments is essential for inductive problems.
\newpage{}

%
%
%
%
%
 \bibliographystyle{splncs04}
 \bibliography{bibfile}

\begin{thebibliography}{10}
\providecommand{\url}[1]{\texttt{#1}}
\providecommand{\urlprefix}{URL }
\providecommand{\doi}[1]{https://doi.org/#1}

\bibitem{holist}
Bansal, K., Loos, S.M., Rabe, M.N., Szegedy, C., Wilcox, S.: {HOL}ist: An
  environment for machine learning of higher order logic theorem proving. In:
  Proceedings of the 36th International Conference on Machine Learning, {ICML}
  2019, 9-15 June 2019, Long Beach, California, {USA}. pp. 454--463 (2019),
  \url{http://proceedings.mlr.press/v97/bansal19a.html}

\bibitem{acl_book}
Boyer, R.S., Moore, J.S.: A computational logic handbook, Perspectives in
  computing, vol.~23. Academic Press (1979)

\bibitem{tactictoe}
Gauthier, T., Kaliszyk, C., Urban, J.: Tactic{T}oe: Learning to reason with
  {HOL4} tactics. In: Eiter, T., Sands, D. (eds.) LPAR-21, 21st International
  Conference on Logic for Programming, Artificial Intelligence and Reasoning,
  Maun, Botswana, May 7-12, 2017. EPiC Series in Computing, vol.~46, pp.
  125--143. EasyChair (2017),
  \url{http://www.easychair.org/publications/paper/340355}

\bibitem{hollight}
Harrison, J.: {HOL} light: {A} tutorial introduction. In: Formal Methods in
  Computer-Aided Design, First International Conference, {FMCAD} '96, Palo
  Alto, California, USA, November 6-8, 1996, Proceedings. pp. 265--269 (1996),
  \url{https://doi.org/10.1007/BFb0031814}

\bibitem{jiang}
Jiang, Y., Papapanagiotou, P., Fleuriot, J.D.: Machine learning for inductive
  theorem proving. In: Artificial Intelligence and Symbolic Computation - 13th
  International Conference, {AISC} 2018, Suzhou, China, September 16-19, 2018,
  Proceedings. pp. 87--103 (2018),
  \url{https://doi.org/10.1007/978-3-319-99957-9\_6}

\bibitem{priority_search_tree}
Lammich, P., Nipkow, T.: Priority search trees. Archive of Formal Proofs  (Jun
  2019), \url{http://isa-afp.org/entries/Priority_Search_Trees.html}, Formal
  proof development

\bibitem{waterfall}
Moore, J.S.: Computational logic : structure sharing and proof of program
  properties. Ph.D. thesis, University of Edinburgh, {UK} (1973),
  \url{http://hdl.handle.net/1842/2245}

\bibitem{acl2}
Moore, J.S.: Symbolic simulation: An {ACL2} approach. In: Formal Methods in
  Computer-Aided Design, Second International Conference, {FMCAD} '98, Palo
  Alto, California, USA, November 4-6, 1998, Proceedings. pp. 334--350 (1998),
  \url{https://doi.org/10.1007/3-540-49519-3\_22}

\bibitem{induction}
Moore, J.S., Wirth, C.: Automation of mathematical induction as part of the
  history of logic. CoRR  \textbf{abs/1309.6226} (2013),
  \url{http://arxiv.org/abs/1309.6226}

\bibitem{Lean}
de~Moura, L.M., Kong, S., Avigad, J., van Doorn, F., von Raumer, J.: The {L}ean
  {T}heorem {P}rover ({S}ystem {D}escription). In: Automated Deduction -
  {CADE-25} - 25th International Conference on Automated Deduction, Berlin,
  Germany, August 1-7, 2015, Proceedings. pp. 378--388 (2015).
  \doi{10.1007/978-3-319-21401-6\_26},
  \url{https://doi.org/10.1007/978-3-319-21401-6\_26}

\bibitem{GitHub}
Nagashima, Y.: data61/psl,
  \url{https://github.com/data61/PSL/releases/tag/v0.1.5-alpha}

\bibitem{meloid}
Nagashima, Y.: Towards machine learning mathematical induction. CoRR
  \textbf{abs/1812.04088} (2018), \url{http://arxiv.org/abs/1812.04088}

\bibitem{lifter}
Nagashima, Y.: {L}i{F}t{E}r: Language to encode induction heuristics for
  {I}sabelle/{HOL}. In: Programming Languages and Systems - 17th Asian
  Symposium, {APLAS} 2019, Nusa Dua, Bali, Indonesia, December 1-4, 2019,
  Proceedings. pp. 266--287 (2019). \doi{10.1007/978-3-030-34175-6\_14},
  \url{https://doi.org/10.1007/978-3-030-34175-6\_14}

\bibitem{evoluaionary_isabelle}
Nagashima, Y.: Towards evolutionary theorem proving for isabelle/hol. In:
  Proceedings of the Genetic and Evolutionary Computation Conference Companion,
  {GECCO} 2019, Prague, Czech Republic, July 13-17, 2019. pp. 419--420 (2019).
  \doi{10.1145/3319619.3321921}, \url{https://doi.org/10.1145/3319619.3321921}

\bibitem{pamper}
Nagashima, Y., He, Y.: Pa{M}pe{R}: proof method recommendation system for
  isabelle/hol. In: Proceedings of the 33rd {ACM/IEEE} International Conference
  on Automated Software Engineering, {ASE} 2018, Montpellier, France, September
  3-7, 2018. pp. 362--372 (2018), \url{https://doi.org/10.1145/3238147.3238210}

\bibitem{psl}
Nagashima, Y., Kumar, R.: A proof strategy language and proof script generation
  for {I}sabelle/{HOL}. In: de~Moura, L. (ed.) Automated Deduction - {CADE} 26
  - 26th International Conference on Automated Deduction, Gothenburg, Sweden,
  August 6-11, 2017, Proceedings. Lecture Notes in Computer Science, vol.
  10395, pp. 528--545. Springer (2017),
  \url{https://doi.org/10.1007/978-3-319-63046-5\_32}

\bibitem{pgt}
Nagashima, Y., Parsert, J.: Goal-oriented conjecturing for isabelle/hol. In:
  Intelligent Computer Mathematics - 11th International Conference, {CICM}
  2018, Hagenberg, Austria, August 13-17, 2018, Proceedings. pp. 225--231
  (2018), \url{https://doi.org/10.1007/978-3-319-96812-4\_19}

\bibitem{concrete_semantics}
Nipkow, T., Klein, G.: Concrete Semantics - With {I}sabelle/{HOL}. Springer
  (2014), \url{https://doi.org/10.1007/978-3-319-10542-0}

\bibitem{isabelle}
Nipkow, T., Paulson, L.C., Wenzel, M.: Isabelle/HOL - a proof assistant for
  higher-order logic, Lecture Notes in Computer Science, vol.~2283. Springer
  (2002)

\bibitem{dfs}
Nishihara, T., Minamide, Y.: Depth first search. Archive of Formal Proofs  (Jun
  2004), \url{http://isa-afp.org/entries/Depth-First-Search.html}, Formal proof
  development

\bibitem{kd_tree}
Rau, M.: Multidimensional binary search trees. Archive of Formal Proofs  (May
  2019), \url{http://isa-afp.org/entries/KD_Tree.html}, Formal proof
  development

\bibitem{hol4}
Slind, K., Norrish, M.: A brief overview of {HOL4}. In: Theorem Proving in
  Higher Order Logics, 21st International Conference, TPHOLs 2008, Montreal,
  Canada, August 18-21, 2008. Proceedings. pp. 28--32 (2008).
  \doi{10.1007/978-3-540-71067-7\_6},
  \url{https://doi.org/10.1007/978-3-540-71067-7\_6}

\bibitem{coq}
{T}he {C}oq~development team: {T}he {C}oq proof assistant,
  \url{https://coq.inria.fr}

\end{thebibliography}
 
 \newpage{}
 
 \section*{Appendix A}\label{s:result}
The three tables in this Appendix give the raw information of the evaluation presented in Section \ref{s:eval}.
The first column ``line'' is the line number in the respective file.
The numbers in the column labelled as ``total'' represent 
how many combinations of arguments \smartinduct{} produced for each inductive problem.
The numbers in the column labelled as ``1st'' show
how many of the created combinations passed the first screening stage explained in Section \ref{s:screening}.
The numbers in the column labelled as ``2nd-b'' show
how many of the created combinations passed the second screening stage explained in Section \ref{s:screening}.
The numbers in the column labelled as ``nth'' show
the ranks \smartinduct{} gave to the combination of induction arguments used by the proof author.
For example, if the number of ``nth'' is 2, this means 
\smartinduct{} recommended the combination of induction arguments
used by the proof author as the second most promising combination.
The numbers in the column labelled as ``score'' represent
the score \smartinduct{} gave to the combination of arguments used by the proof author.
For example, if the number of ``score'' is 18, this means
\smartinduct{} gave 18 points to the combination of argument used by the proof author,
indicating that two feature extractors wrongly judged that the combination is not appropriate for
the proof goal under consideration,
assuming that the human proof author is
always right.
The values in the column labelled as \verb|rule| show
whether the proof author passed an argument to the \verb|rule| field.
The values in the column labelled as \verb|arb| show
whether the proof author passed an argument to the \verb|arbitrary| field.

\begin{table}[H]
\caption{Evaluation of \smartinduct{} on \texttt{DFS.thy}}
\label{table:dfs}
\begin{center}
\begin{tabular}{r c r r r r r r c c}
\hline\noalign{\smallskip}
line & theorem name & total & 1st & 2nd-a & 2nd-b & nth & score & \verb|rule| & \verb|arb| \\
\hline
27  & \verb|nexts_set| & 128 & 16 & 12 & 12 & 2 & 20 & no & no                                 \\
42  & \verb|-| & 20 & 8 & 4 & 4 & 2 & 20 & no & no                                             \\
87  & \verb|df2_invariant| & 256 & 40 & 32 & 32 & - & - & yes & no                             \\
126 & \verb|dfs_app| & 3210 & 590 & 576 & 576 & 1 & 20 & yes & no                              \\
131 & \verb|-| & 384 & 144 & 136 & 136 & 2 & 20 & yes & no                                     \\
137 & \verb|visit_subset_dfs| & 384 & 144 & 136 & 136 & 1 & 20 & yes & no                      \\
140 & \verb|next_subset_dfs| & 384 & 144 & 136 & 136 & 1 & 20 & yes & no                       \\
161 & \verb|nextss_closed_dfs’| & 384 & 144 & 136 & 136 & 1 & 20 & yes & no                    \\
176 & \verb|Image_closed_trancl| & 20 & 8 & 8 & 8 & 1 & 18 & no & no                           \\
206 & \verb|dfs_subset_reachable...| & 384 & 144 & 136 & 136 & 1 & 20 & yes & no               \\
\hline
\end{tabular}
\end{center}
\end{table}

\setlength{\tabcolsep}{3pt}
\begin{table}[h]
\caption{Evaluation of \smartinduct{} on \texttt{RST\_RBT.thy}}
\label{table:rst}
\begin{center}
\begin{tabular}{r c r r r r r r c c}
\hline\noalign{\smallskip}
line & theorem name & total & 1st & 2nd-a & 2nd-b & nth & score & \verb|rule| & \verb|arb| \\
\hline
170 & \verb|inorder_combine| & 60 & 44 & 40 & 40 & 1 & 20 & yes & no                          \\
175 & \verb|inorder_upd| & 640 & 80 & 72 & 64 & 1 & 20 & yes & no                             \\
185 & \verb|inorder_del| & 100 & 36 & 32 & 28 & 1 & 20 & yes & no                             \\
227 & \verb|inv_baliL| & 512 & 136 & 128 & 128 & 1 & 20 & yes & no                            \\
232 & \verb|invc_baliR| & 512 & 136 & 128 & 128 & 1 & 20 & yes & no                           \\
242 & \verb|bheight_baliL| & 384 & 112 & 104 & 104 & 1 & 20 & yes & no                        \\
247 & \verb|bheight_baliR| & 384 & 112 & 104 & 104 & 1 & 20 & yes & no                        \\
257 & \verb|invh_mkNode| & 512 & 136 & 128 & 128 & 1 & 20 & yes & no                          \\
262 & \verb|invh_baliR| & 512 & 136 & 128 & 128 & 1 & 20 & yes & no                           \\
269 & \verb|invc_upd| & 640 & 96 & 96 & 96 & 1 & 20 & yes & no                                \\
276 & \verb|invh_upd| & 384 & 64 & 56 & 56 & 1 & 20 & yes & no                                \\
291 & \verb|invpst_upd| & 384 & 64 & 56 & 56 & 1 & 20 & yes & no                              \\
309 & \verb|invh_baldL_invc| & 768 & 184 & 176 & 176 & 1 & 20 & yes & no                      \\
316 & \verb|invh_baldL_Black| & 640 & 160 & 160 & 160 & 1 & 20 & yes & no                     \\
320 & \verb|invc_baldL| & 640 & 160 & 160 & 160 & 1 & 20 & yes & no                           \\
324 & \verb|invc2_baldL| & 512 & 136 & 128 & 128 & 1 & 20 & yes & no                          \\
331 & \verb|invh_baldR_invc| & 768 & 184 & 176 & 176 & 1 & 20 & yes & no                      \\
336 & \verb|inv_baldR| & 640 & 160 & 160 & 160 & 1 & 20 & yes & no                            \\
340 & \verb|inv2_baldR| & 512 & 136 & 128 & 128 & 1 & 20 & yes & no                           \\
347 & \verb|invh_combine| & 80 & 56 & 52 & 52 & 1 & 20 & yes & no                             \\
356 & \verb|inv_combine| & 100 & 68 & 68 & 68 & 1 & 20 & yes & no                             \\
366 & \verb|del_inv_invh| & 140 & 60 & 60 & 60 & 1 & 20 & yes & no                            \\
398 & \verb|invpst_combine| & 60 & 44 & 40 & 40 & 1 & 20 & yes & no                           \\
403 & \verb|invpst_del| & 60 & 28 & 24 & 24 & 1 & 20 & yes & no                               \\
\hline
\end{tabular}
\end{center}
\end{table}

\begin{table}[H]
\caption{Evaluation of \smartinduct{} on \texttt{Nearest\_Neighbors.thy}}
\label{table:nearest_neibors}
\begin{center}
\begin{tabular}{r c r r r r r r c c}
\hline\noalign{\smallskip}
line & theorem name & total & 1st & 2nd-a & 2nd-b & nth & score & \verb|rule| & \verb|arb| \\
\hline
66  & \verb|sqed_ge_0| & 40 & 32 & 27 & 27 & 1 & 20 & yes & no                     \\
71  & \verb|sqed_eq_0| & 40 & 32 & 27 & 27 & 1 & 20 & yes & no                     \\
76  & \verb|sqed_eq_0_rev| & 40 & 32 & 27 & 27 & 1 & 20 & yes & no                 \\
81  & \verb|sqed_com| & 40 & 32 & 27 & 27 & 2 & 20 & yes & no                      \\
147 & \verb|minimize_sqed| & 256 & 72 & 62 & 62 & - & - & yes & yes                \\
228 & \verb|sorted_insort_sqed| & 128 & 32 & 32 & 32 & 8 & 18 & no & no            \\
285 & \verb|sorted_sqed_last_take_mono| & 256 & 48 & 44 & 44 & 5 & 20 & no & yes   \\
292 & \verb|sorted_sqed_last_insort_eq| & 128 & 32 & 32 & 32 & 4 & 20 & no & no    \\
321 & \verb|mnn_length| & 3210 & 192 & 192 & 192 & - & - & no & yes                \\
328 & \verb|mnn_length_gt_0| & 2080 & 160 & 160 & 160 & 7 & 20 & no & yes          \\
335 & \verb|mnn_length_gt_eq_m| & 2080 & 119 & 119 & 119 & 7 & 20 & no & yes       \\
344 & \verb|mnn_sorted| & 2080 & 160 & 160 & 160 & 5 & 20 & no & yes               \\
350 & \verb|mnn_set| & 3210 & 192 & 192 & 192 & - & - & no & yes                   \\
359 & \verb|mnn_distinct| & 2080 & 160 & 160 & 160 & 7 & 20 & no & yes             \\
388 & \verb|mnn_le_last_ms| & 3120 & 192 & 184 & 184 & 7 & 20 & no & yes           \\
431 & \verb|mnn_sqed| & 4160 & 204 & 188 & 188 & 7 & 20 & no & yes                 \\
\hline
\end{tabular}
\end{center}
\end{table}

\newpage{}
\section*{Appendix B}\label{app:screenshot}
Figure \ref{fig:smartinduct} illustrates the user-interface of \smartinduct{}.

\begin{figure}[H]
\centering
  \includegraphics[width=1.0\textwidth]{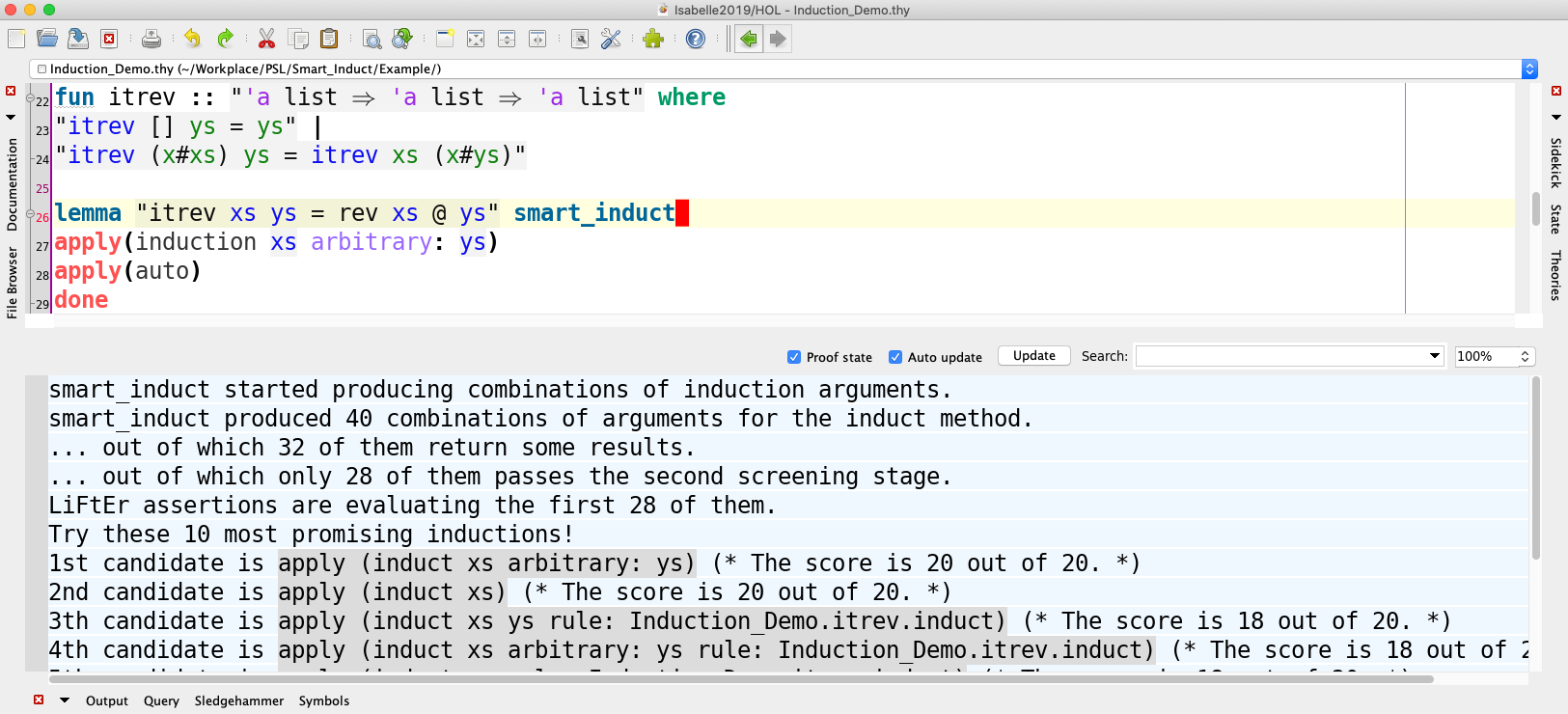}
  \caption{A Screenshot of Isabelle/jEdit with \smartinduct.}
  \label{fig:smartinduct}
\end{figure}
\end{document}